\documentclass[runningheads]{llncs}
\usepackage{graphicx}
\usepackage{multirow, multicol}
\usepackage[table,xcdraw]{xcolor}
\usepackage{colortbl}
\definecolor{Green}{rgb}{0.196, 0.854, 0.186}
\definecolor{Yellow}{rgb}{1.0, 1.0, 0.0}
\usepackage{tikz}
\usepackage{comment}
\usepackage{amsmath,amssymb} % define this before the line numbering.
\usepackage{color}
\usepackage{pifont}

\usepackage[accsupp]{axessibility}  % Improves PDF readability for those with disabilities.

\begin{document}
% \linenumbers
\pagestyle{headings}
\mainmatter
\def\ECCVSubNumber{3966}  % Insert your submission number here

\title{DeepPS2: Revisiting Photometric Stereo using Two Differently Illuminated Images} % Replace with your title

\begin{comment}
\titlerunning{ECCV-22 submission ID \ECCVSubNumber} 
\authorrunning{ECCV-22 submission ID \ECCVSubNumber} 
\author{Anonymous ECCV submission}
\institute{Paper ID \ECCVSubNumber}
\end{comment}
%******************

% \begin{comment}
\titlerunning{DeepPS2}
\author{Ashish Tiwari\inst{1}\orcidID{0000-0002-4462-6086} \and
Shanmuganathan Raman\inst{2}\orcidID{0000-0003-2718-7891} }
\authorrunning{A. Tiwari and S. Raman}
\institute{Prime Minister Research Fellow \and 
Jibaben Patel Chair in Artificial Intelligence\\CVIG Lab, IIT Gandhinagar, Gujarat, India
\email{\{ashish.tiwari,shanmuga\}@iitgn.ac.in}}
% % \end{comment}
%******************
\maketitle

\begin{abstract}
Photometric stereo, a problem of recovering 3D surface normals using images of an object captured under different lightings, has been of great interest and importance in computer vision research. Despite the success of existing traditional and deep learning-based methods, it is still challenging due to: (i) the requirement of three or more differently illuminated images, (ii) the inability to model unknown general reflectance, and (iii) the requirement of accurate 3D ground truth surface normals and known lighting information for training. In this work, we attempt to address an under-explored problem of photometric stereo using just two differently illuminated images, referred to as the PS2 problem. It is an intermediate case between a single image-based reconstruction method like Shape from Shading (SfS) and the traditional Photometric Stereo (PS), which requires three or more images. We propose an inverse rendering-based deep learning framework, called DeepPS2, that jointly performs surface normal, albedo, lighting estimation, and image relighting in a completely self-supervised manner with no requirement of ground truth data. We demonstrate how image relighting in conjunction with image reconstruction enhances the lighting estimation in a self-supervised setting.\footnote{Supported by SERB IMPRINT 2 Grant}

\end{abstract}

\section{Introduction}
Inferring the 3D shape of the objects using digital images is a fundamental and challenging task in computer vision research. It directly extends to quality control, virtual/augmented reality, medical diagnosis, e-commerce, etc. The widely used geometric approaches to shape recovery such as binocular \cite{kendall2017end,taniai2017continuous} or multi-view stereo \cite{schonberger2016structure,furukawa2009accurate,kumar2017monocular,kumar2019jumping,kumar2019superpixel} methods require images from different viewpoints to triangulate the 3D points. However, they rely heavily on the success of image feature matching techniques and fall short of recovering finer details such as indentations, imprints, and scratches. The photometric methods for 3D shape reconstruction use shading cues from either a single image - \textit{Shape from Shading (SfS)} \cite{horn1970shape} or at least three images - \textit{Photometric Stereo (PS)} \cite{woodham1980photometric} to recover surface normals and are known to better preserve the finer surface details. 

\textbf{What are the bottlenecks?} The SfS problem is ill-posed due to the underlying convex/concave ambiguity and the fact that infinite surface normals exist to explain the intensity at each pixel \cite{prados2005shape}. The PS methods are known to handle such ambiguities and provide a unique surface normal defining the intensity at each pixel by using three or more differently illuminated images. However, the well-posed traditional photometric stereo problem (as introduced by Woodhman \cite{woodham1980photometric}) assumes the surfaces to be purely Lambertian, which seldom is the case in the real world. Several recent methods \cite{ikehata2018cnn,zheng2019spline,chen2018ps,chen2020deep,chen2019self,chen2020learned} have also addressed shape estimation for non-Lambertian surfaces with unknown reflectance properties. However, they require more images ($\sim 50-100$) as input. While there are methods that require as few as six (or even fewer) images \cite{lichy2021shape}, our goal is to resort to just two images under a photometric stereo setting, referred to as a PS2 problem.

\textbf{Scope of the PS2 problem}. The scope of this work is to address the photometric stereo problem in an intermediate setting with two images ($m=2$) between SfS ($m=1$) and the traditional PS ($m \geq 3$). It can essentially be viewed as a degenerative case of lack of meaningful information due to shadows in a typical three-source photometric stereo setting \cite{hernandez2010overcoming}. Another use case of a PS2 problem arises in the 3D reconstruction of the non-rigid objects \cite{hernandez2007non}. When an object is imaged under three light sources, one could be occluded by the object, and only the other two would provide meaningful cues. Therefore, a PS2 problem needs to be solved in such cases. Further, the PS2 problem arises when $m \geq 3$ and light sources are coplanar. Such a situation typically occurs when the scene is illuminated by the sun and hence, applies to outdoor PS as well \cite{sato1995reflectance,jung2015one}. 

\textbf{Constraints in addressing the PS2 problem.} There exists two formulations of photometric stereo, in general -  the \textit{differential} and the \textit{non-differential} formulation. Several normal fields can offer solutions to the PS2 problem. Under either of the settings, a remedy is to perform an exhaustive search among these normal fields and smoothly find the best one that characterizes the underlying shape. In other words, the task is to find the normal field that best satisfies the smoothness constraint \cite{onn1990integrability}. The differential approach of PS implicitly enforces such smoothness. However, it requires explicit knowledge of the surface boundary conditions \cite{mecca2011unambiguous}, which is rarely available or requires regularization \cite{hernandez2010overcoming}, which is generally tedious owing to heavy parameter tuning. A few methods \cite{mecca2011unambiguous,queau2017photometric} have put forward ways to address the PS2 problem based on the non-differential formulation by recasting it as a binary labeling problem. While such optimization problems can be solved using graph-cut-based algorithms \cite{boykov2001fast}, they require the albedo to be known. 

\textbf{Can deep neural networks offer a solution?} Owing to its success and applicability in solving the general PS problem, we intend to address the PS2 problem using deep neural networks. The core idea is to use a deep neural network to model unknown general surfaces with complex Bidirectional Reflectance Distribution Functions (BRDFs). The photometric stereo problem using deep neural networks has been addressed either under a \textit{calibrated} (known lightings) or an \textit{uncalibrated} (unknown lightings) setting. While most of these methods require 3D ground truth supervision \cite{ikehata2018cnn,zheng2019spline,chen2018ps,chen2020deep,chen2019self,chen2020learned,li2019learning,taniai2018neural}, a little progress has been made to address PS in a self-supervised manner \cite{kaya2021uncalibrated}. However, such self-supervised and uncalibrated methods still require ground truth supervision for lighting estimation. 

In this work, we introduce an inverse-rendering-based deep learning framework, called DeepPS2, to address the PS2 problem and work towards developing a completely uncalibrated and self-supervised method. The core idea is to utilize the shading cues from two differently illuminated images to obtain the 3D surface normals. DeepPS2 is designed to perform albedo estimation, lighting estimation, image relighting, and image reconstruction without any supervision using 3D surface normals and/or lighting. While image reconstruction is commonly adopted in the existing unsupervised/self-supervised approaches, the appropriate design considerations to perform image relighting using the estimated lightings bring out several interesting insights about the proposed framework.   

\textbf{Contributions} The following are the key contributions of this work.
\begin{itemize}
    \item We introduce DeepPS2, an uncalibrated deep learning-based photometric stereo method that jointly performs surface normal, albedo, and lighting estimation in a self-supervised setting. The workflow of the proposed framework follows the principles of inverse-rendering.
    \item We propose a self-supervised lighting estimation through light space discretization and inclusion of image relighting (using the estimated lightings) along with image reconstruction.
    \item We propose to explicitly model the specularities through albedo refinement and estimated illumination.
    \item To the best of our knowledge, ours is the first work to address the PS2 problem in a deep learning setting and lighting estimation in a self-supervised setting for the task at hand.
\end{itemize}

\section{Related Work}
\label{sec:related_work}
This section reviews the literature on the PS2 problem and some recent deep learning-based photometric stereo methods.

\textbf{The PS2 Problem.} Some earlier works have addressed this problem in a traditional non-learning setting. Onn and Bruckstein \cite{onn1990integrability} discussed the ambiguities in determining surface normals using two images and proposed to use integrability constraint to handle such ambiguities. Sato and Ikeuchi \cite{sato1995reflectance} used their method to solve the problem with $m \geq 3$ images under solar illumination, which in a sense addresses the PS2 problem \cite{woodham1980photometric}. Later, Yang \emph{et al.} \cite{yang1992two} studied the problem, particularly for the convex objects. Kozera provided an analytical resolution to the differential formulation of PS2 \cite{kozera1992shape}. Since 1995 (for over ten years later), only Ikeda \cite{ikeda2003robust} addressed the PS2 problem by
essentially considering the second image as an auxiliary to better solve the SFS problem. Recently, Queau \emph{et al.} \cite{queau2017photometric} addressed the PS2 problem using a graph cut based optimization method. Further, the problem of outdoor PS is being re-explored in several works \cite{abrams2012heliometric,ackermann2012photometric}. While these methods attempt to provide a numerical resolution to the PS problem \cite{mecca2011unambiguous,queau2017photometric}, we intend to address it using the capacity of deep neural networks. 

\textbf{Deep Learning-based methods.} Deep learning has seen great progress in many areas of computer vision, including photometric stereo \cite{zheng2019spline,chen2018ps,chen2019self,chen2020learned,ikehata2018cnn,santo2017deep}. Santo \emph{et al.} \cite{santo2017deep} were the first to propose a deep learning-based method to obtain per-pixel surface normals. However, they were limited by the pre-defined order of pixels at the input. Later, Chen \emph{et al.} in their subsequent works \cite{chen2018ps,chen2019self,chen2020learned} proposed to model the spatial information using feature-extractor and features-pooling based strategies for photometric stereo. Further, the works by Yao \emph{et al.} \cite{yao2020gps} and Wang \emph{et al.} \cite{wang2020non} proposed to extract and combine the local and global features for better photometric understanding. However, all these methods require ground truth surface normals for supervision which is generally difficult to obtain. Recently, Taniai \& Maehara \cite{taniai2018neural} proposed a self-supervised network to directly output the surface normal using a set of images and reconstruct them. However, their method required known lightings as input. Kaya \emph{et al.} \cite{kaya2021uncalibrated} expanded their method to deal with inter-reflections and address photometric stereo in an uncalibrated setting. However, the lighting estimation was still performed using ground truth supervision. Other methods such as Lichy \emph{et al.} \cite{lichy2021shape}, and Boss \emph{et al.} \cite{boss2020two} predicted shape and material using three or less and two images (one with and one without flash), respectively. While LERPS \cite{tiwari2022lerps} infers lighting and surface normal from a single image, it requires multiple images (one at a time) for training. We work towards an uncalibrated photometric stereo method that uses only two differently illuminated images as the input while estimating lightings, surface normals, and albedos, all in a self-supervised manner.   

\section{Understanding PS2: Photometric Stereo using Two Images}
Before describing the PS2 problem that we are interested to address, we would like to review some key features of the SfS  \cite{horn1970shape} and the traditional PS problem \cite{woodham1980photometric}. We assume that an orthographic camera images the surface under uniform directional lighting with viewing direction $\boldsymbol{v} \in \rm I\!R^{3}$ pointing along the z-direction and the image plane parallel to the $XY$ plane of the 3D Cartesian coordinate system $XYZ$.

\subsection{Shape from Shading (SfS)}
Consider an anisotropic non-Lambertian surface $f$ characterised by the Bidirectional Reflectance Distribution Function (BRDF) $\rho$.  For each surface point $(x,y)$ characterized by the surface normal $\boldsymbol{n} \in \rm I\!R^{3}$ and illuminated the light source in the direction $\boldsymbol{\ell} \in \rm I\!R^{3}$, the image formation of the surface viewed from the direction $\boldsymbol{v} \in \rm I\!R^{3}$ is given by Equation \ref{img_form_sfs}.
    \begin{equation}
        \boldsymbol{I}(x,y) = \rho(\boldsymbol{n},\boldsymbol{\ell},\boldsymbol{v})\psi_{f,s}(x,y) \left[\boldsymbol{n}(x,y)^{T}\boldsymbol{\ell}\right] + \epsilon
    \label{img_form_sfs}
    \end{equation}
Here, $\psi_{f,s}(x,y)$ specifies the attached and the cast shadows. It is equal to 0, if $(x,y)$ is shadowed and equal to 1, otherwise. $\epsilon$ incorporates the global illumination and noise effect. $\boldsymbol{I}(x,y)$ is the normalized gray level with respect to the light source intensity. 

Clearly, with albedo and lightings being known apriori, the surface normals $\boldsymbol{n}(x,y)$ in the revolution cone around the lighting direction $\boldsymbol{\ell}$ constitute the set of infinite solutions to Equation \ref{img_form_sfs}. Therefore, it becomes an ill-posed problem and is difficult to solve locally.

\subsection{Photometric Stereo (PS)}
The simplest solution to overcome the ill-posedness of SfS is to have $m \geq 2$ differently illuminated images of the object taken from the same viewpoint. In general, for multiple light sources, the formulation described in Equation \ref{img_form_sfs} extends to the following.
\begin{equation}
        \boldsymbol{I}_{j}(x,y) = \rho(\boldsymbol{n},\boldsymbol{\ell}_{j},\boldsymbol{v})\psi_{f,s}(x,y) \left[\boldsymbol{n}(x,y)^{T}\boldsymbol{\ell}_{j}\right] + \epsilon_{j}
\label{img_form_ps}
\end{equation}
Here, the equation is specific to the $j^{th}$ light source. For $m \geq 3$ and a Lambertian surface, Equation \ref{img_form_ps} formulates a photometric stereo problem (the traditional one for $m = 3$). Solving such a system is advantageous as it is well-posed and can be solved locally, unlike SfS. 

\subsection{The PS2 problem}
With such a non-differential formulation (as in Equation \ref{img_form_ps}), the three unknowns  $(n_x, n_y, n_z)$ can be obtained by solving three or more linear equations. However, such a formulation is tricky to solve under two scenarios: (i) when the light sources are coplanar (rank-deficit formulation) and (ii) when $m = 2$. These scenarios lead us to the formulation of the PS2 problem - photometric stereo with two images, as described in Equation \ref{img_form_ps2}.
\begin{equation*}
        \rho(\boldsymbol{n},\boldsymbol{\ell}_{1},\boldsymbol{v})\psi_{f,s}(x,y) \left[\boldsymbol{n}(x,y)^{T}\boldsymbol{\ell}_{1}\right] + \epsilon_{1} = \boldsymbol{I}_{1}(x,y) 
\end{equation*}
\begin{equation*}
         \rho(\boldsymbol{n},\boldsymbol{\ell}_{2},\boldsymbol{v})\psi_{f,s}(x,y) \left[\boldsymbol{n}(x,y)^{T}\boldsymbol{\ell}_{2}\right] + \epsilon_{2} =\boldsymbol{I}_{2}(x,y)
\end{equation*}
\begin{equation}
        n_{x}(x,y)^{2} +  n_{y}(x,y)^{2} +  n_{z}(x,y)^{2} = 1
        \label{img_form_ps2}
\end{equation}
The non-linearity in the third part of Equation \ref{img_form_ps2} could give non-unique solution \cite{ikeuchi1981numerical}. Adding one more image (under non-coplanar light source configuration) can straightaway solve the problem. However, it will fail when the surface is arbitrarily complex in its reflectance properties. Further, the problem becomes even more difficult to solve when albedo is unknown.

To address the aforementioned issues in the PS2 problem with unknown albedo and lightings, we introduce a deep learning-based framework that can resolve such ambiguities by directly learning from images.

\section{Method}
In this section, we describe DeepPS2, a deep learning-based solution to the PS2 problem. Further, we describe several design considerations, light space sampling and discretization, and share the training strategy.

\begin{figure}
\centering
\includegraphics[width=\linewidth]{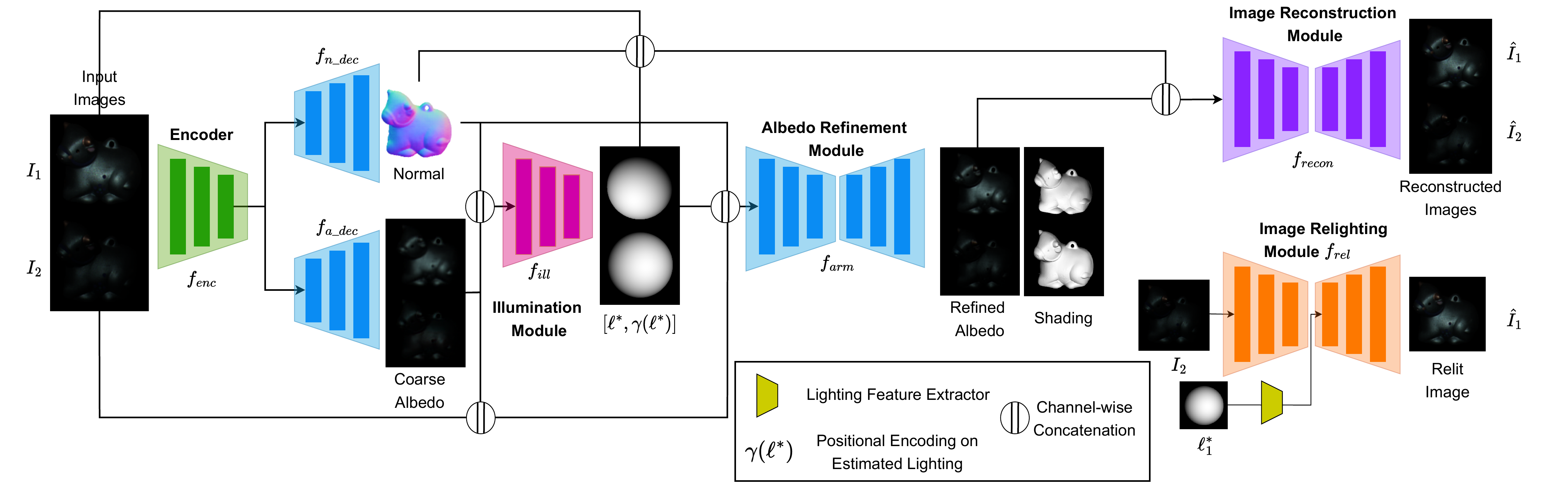}
\caption{The proposed inverse rendering framework, called DeepPS2, for shape, material, and illumination estimation. The encoder-decoder design is inspired by Hourglass networks \cite{yang2017stacked}. Layer-wise skip connections are avoided for visual clarity}
\label{fig:block_diag}
\end{figure}
\vspace{-1cm}

\subsection{Network Architecture}
Let $I_{1}, I_{2} \in \rm I\!R^{C \times H \times W} $ be the two images corresponding to the lighting directions $\boldsymbol{\ell}_{1}$ and $\boldsymbol{\ell}_{2}$, respectively. The two images along with the object mask $M \in \rm I\!R^{1 \times H \times W}$ are fed to the encoder $f_{enc}$ to obtain an abstract feature map $\boldsymbol{\phi}_{img}$, as described in Equation \ref{eq:enc_feat}.
\begin{equation}
    \centering
    \boldsymbol{\phi}_{img} = f_{enc}([I_{1}, I_{2}, M]; \boldsymbol{\theta}_{enc})
    \label{eq:enc_feat}
\end{equation}
Here, $[\cdot]$ represents channel-wise concatenation and $\boldsymbol{\theta}_{enc}$ represents the parameters of the encoder.

\textbf{Surface Normal and Albedo Estimation.} We use $\boldsymbol{\phi}_{img}$ to obtain an estimate of surface normal map $\hat{N}$ and the albedo $\hat{A}$ through the decoders $f_{n\_dec}$ and $f_{a\_dec}$, respectively, as described in Equation \ref{eq:norm_alb_dec}.
\begin{equation*}
    \centering
    \hat{N} = f_{n\_dec}(\boldsymbol{\phi}_{img}; \boldsymbol{\theta}_{n\_dec}) 
\end{equation*}
\begin{equation}
    \centering
    \hat{A} = f_{a\_dec}(\boldsymbol{\phi}_{img}; \boldsymbol{\theta}_{a\_dec})
    \label{eq:norm_alb_dec}
\end{equation}
Here, $\hat{A} = [\hat{A}_{1}, \hat{A}_{2}]$ represents the albedos of two images $I_{1}$ and $I_{2}$ together. The design of each encoder-decoder combination \footnote{The detailed layer-wise architecture can be found in our supplementary material.} is inspired by that of the Hourglass network \cite{yang2017stacked}.

\textbf{Lighting Estimation.} A straightforward way to estimate lighting directions could be to use another fully connected branch and train the network to regress to the desired lightings directly from $\boldsymbol{\phi}_{img}$. However, fully connected layers require a large number of parameters. Further, obtaining precise lighting information directly just from the image features would be difficult since it would not have the explicit knowledge of the structure and reflectance properties of the underlying surface. With an intent to keep the entire architecture fully convolutional, we propose an \textit{illumination module} ($f_{ill}$) to predict the desired lighting directions by using the estimated normal map and albedos, as described in Equation \ref{eq:lem}.
\begin{equation}
    \centering
    \hat{l_{i}} = f_{ill}([\hat{N}, \hat{A}_{i}];\boldsymbol{\theta}_{lem})
    \label{eq:lem}
\end{equation}
Here, $i=1,2$ corresponding to two images $I_{1}$ and $I_{2}$, respectively.

At this stage, one straightforward approach could be to use the estimated normal, albedos, and lightings in order to reconstruct the original images through the image rendering equation (see Equation \ref{eq:recon}). However, the estimated albedo $\hat{A}$ without lighting estimates fails to capture the complex specularities on the surface (see Figure \ref{fig:res_rel}). Also, the estimated lightings were a little far from the desired ones. 

Thus, the question now is - \textit{how do we validate the accuracy of the estimated albedos and lightings}, especially when there is no ground truth supervision? The albedos and lightings go hand-in-hand and are dependent on each other as far as image rendering is considered, of course, in addition to the surface normal (see Generalized Bas Relief (GBR) ambiguity \cite{belhumeur1999bas}). To address the aforementioned concerns, we propose two crucial resolves: (i) \textit{albedo refinement} before image reconstruction and (ii) \textit{image relighting} using the estimated lightings.

\textbf{Albedo Refinement by Specularity Modeling.} As discussed earlier, the estimated albedo $\hat{A}$ failed to represent the specularities directly from the image features. Most of the existing deep photometric stereo methods have implicitly handled specularities using multiple differently illuminated images through max-pooling and global-local feature-fusion. However, it is crucial to understand that the specularities are essentially the reflections on the surface, and information about surface geometry can help model such specularities better. Understanding surface geometry becomes even more crucial when we have just one or two images to model the surface reflection. Therefore, we choose to explicitly model these specularities and refine the albedo estimate using a few reasonable and realistic assumptions. 

We assume that the specular BRDF is isotropic and is only the function of the half-vector $\boldsymbol{h}$ and the surface normal $\boldsymbol{n}$ at any point on the surface as the BRDF can be re-parameterized to a half-vector based function \cite{rusinkiewicz1998new}. In doing so, we could omit the Fresnel Reflection coefficients and geometric attenuation associated with modelling BRDFs. The authors in \cite{pacanowski2012rational,burley2012physically} found that the isotropic BRDF can also be modeled simply by two parameters $\theta_{h} = cos^{-1}(\boldsymbol{n}^{T}\boldsymbol{h})$ and $\theta_{d}= cos^{-1}(\boldsymbol{v}^{T}\boldsymbol{h})$. Therefore, we use the estimated lighting $\ell_{i}$ to compute $cos(\theta_{h})$ and $cos(\theta_{d})$ to further refine the albedo. Additionally,  we use positional encoding to model the high-frequency specularities in the refined albedo. In short, we construct the $L_{i}$ as per Equation \ref{eq:light}.
\begin{equation*}
    \centering
    L_{i} = [\boldsymbol{p}_{i}, \gamma(\boldsymbol{p}_{i})]
\end{equation*}
\begin{equation}
    \centering
    \boldsymbol{p}_{i} = [\boldsymbol{n}^{T}\boldsymbol{h}_{i}, \boldsymbol{v}^{T}\boldsymbol{h}_{i}]
    \label{eq:light}
\end{equation}
Here, $\gamma(\eta) = [sin(2^{0}\pi\eta), cos(2^{0}\pi\eta), ... , sin(2^{m-1}\pi\eta), cos(2^{m-1}\pi\eta)]$. We choose $m=3$ in our method. Futher, $\boldsymbol{h}_{i} = \frac{\hat{\boldsymbol{l}_{i}} + \boldsymbol{v}}{||\hat{\boldsymbol{l}_{i}} + \boldsymbol{v}||}$.

Following these observations, we use an encoder-decoder based \textit{albedo refinement module} ($f_{arm}$) to obtain the refined albedo by considering the estimated lightings $L_{i}$, albedos $\hat{A}$, surface normal $\hat{N}$, and the underlying images as its input. Equation \ref{eq:arm} describes the information flow.
\begin{equation}
    \centering
    \hat{A}_{i(ref)} = f_{arm}([I_{i}, \hat{N}, \hat{A}_{i}, L_{i},]; \boldsymbol{\theta}_{arm})
    \label{eq:arm}
\end{equation}

\textbf{Image Relighting.} Generally, at this stage, the existing approaches proceed further to use the rendering equation and reconstruct the input image(s). However, the lightings are either known or have been estimated with ground truth supervision. This allows stable training and offers convincing results. However, in our case, the lightings are estimated without any explicit supervision and are expected to produce learning instabilities. So the question is, \textit{how can we ensure that the estimated lightings are close to the desired ones without any ground truth supervision?} 

As an additional check on the authenticity of the estimated lightings, we propose to use them for the image relighting task. We use an \textit{image relighting module} $(f_{rel})$ to relight one image into the other using the estimated lighting as the target lighting and measure the quality of the relit image, as described in Equation \ref{eq:relight}.
\begin{equation}
    \centering
    \hat{I}_{1(rel)} = f_{rel}(I_{2},\boldsymbol{\phi}(\hat{\boldsymbol{\ell}_{1}}); \boldsymbol{\theta}_{rel})
    \label{eq:relight}
\end{equation}
Here, $\boldsymbol{\phi}(\hat{\boldsymbol{\ell}_{1}})$ is the lighting feature extracted from the desired target lighting $\hat{\boldsymbol{\ell}_{1}}$. The quality of the relit image fosters the lighting estimates to be close to the desired ones.

\textbf{Image Reconstruction.} Having obtained the estimates of surface normal, albedo, and lightings, we finally use them to obtain the reflectance map $\boldsymbol{R}_{i}$ using the encoder-decoder based \textit{image reconstruction module} $(f_{recon})$, as described in Equation \ref{eq:reflect}.
\begin{equation}
    \centering
    \boldsymbol{R}_{i} = f_{recon}([I_{i}, \hat{N}, \hat{A}_{i(ref)}, \hat{\boldsymbol{\ell}_{i}}]; \boldsymbol{\theta}_{recon})
    \label{eq:reflect}
\end{equation}

The reflectance image $\boldsymbol{R}_{i}$ is then used to reconstruct the associated image $\hat{I}_{i}$, as described in Equation \ref{eq:recon}.
\begin{equation}
    \centering
    \hat{I}_{i} = \boldsymbol{R}_{i} \odot max(\hat{\boldsymbol{\ell}_{i}}^{T}\hat{N}, 0)
    \label{eq:recon}
\end{equation}
Here, $\odot$ refers to the element-wise multiplication.

In this way, the proposed DeepPS2 produces estimates of surface normal, albedos, and lightings as well as relights the image under target lightings by using only two images as input and no additional ground truth supervision. Based on the network performance, we show that the PS2 problem can be well addressed using the benefits of deep learning framework. 
\vspace{-0.5cm}
\begin{figure}
\centering
\includegraphics[width=0.9\linewidth]{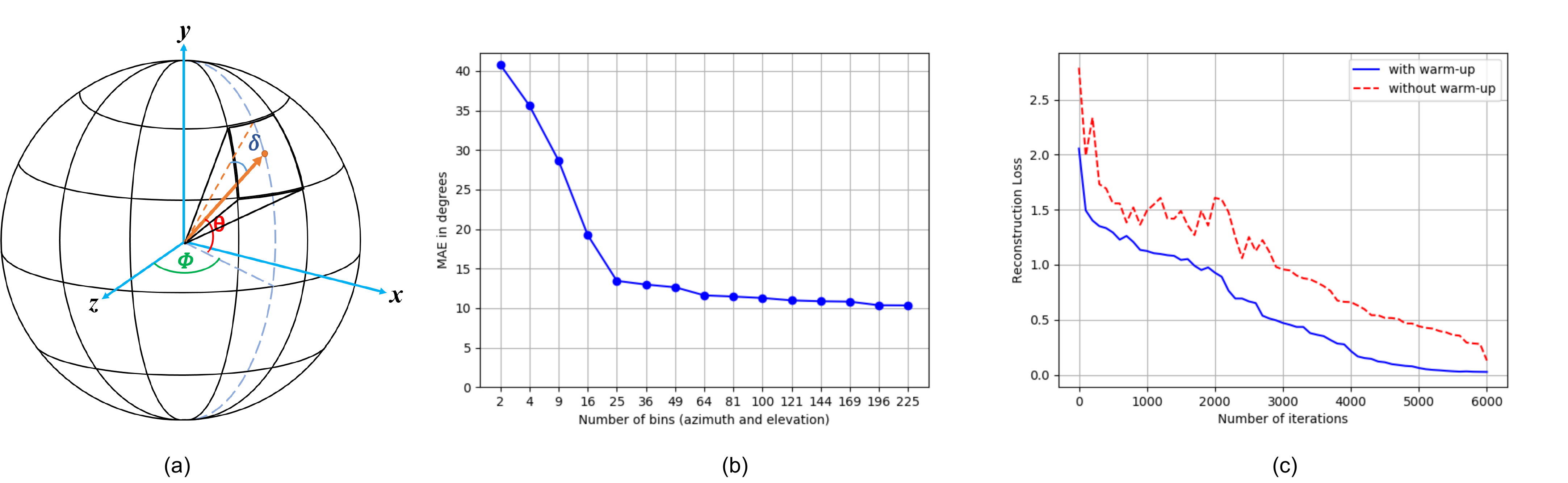}
\caption{(a) Light space discretization into $K=25$ bins. $\delta = 180/2K$ is the maximum angular deviation. (b) Variation of MAE with $K$. (c) Effect of early stage warm-up}
\label{fig:light_space}
\end{figure}
\vspace{-1cm}
\subsection{More on Lighting Estimation: The Light Space Sampling}
As discussed earlier, an intuitive approach to estimate light source directions would be to directly regress them from image(s). However, regressing these values to the exact ones is difficult and can cause learning difficulties \cite{chen2019self}. Further, under the distant light source assumption, it is easier and better to specify a region in the light space rather than the exact direction while locating the light source. Additionally, this eases the light source calibration during data acquisition. Therefore, we choose to formulate the lighting estimation as a classification problem. A few methods in the recent past have adopted the classification formulation \cite{chen2019self,chen2020learned} and weak calibration setting \cite{lichy2021shape} for lighting and shape estimation and have produced excellent results. 

In this work, we discretize the light space (upper hemisphere) into $K=25$ bins (as shown in Fig. \ref{fig:light_space}(a)) i.e. 5 bins along the azimuth direction $\phi \in [0^{\circ}, 180^{\circ}]$ centered at $[18^{\circ}, 54^{\circ}, 90^{\circ}, $ $126^{\circ}, 162^{\circ}]$ and $5$ bins along the elevation direction $\theta \in [-90^{\circ}, 90^{\circ}]$ centered at $[-72^{\circ}, -36^{\circ}, 0^{\circ}, 36^{\circ}, 72^{\circ}]$. While each bin suffers a maximum angular deviation of $18^{\circ}$ along each direction (Fig. \ref{fig:light_space}(a)), they offer a relatively simpler light source configuration during data acquisition. They can be realized using hand-held lighting devices. Further, learning under such discretized light space configuration allows the network to better tolerate errors in the estimated lightings and the subsequent downstream tasks. During training, the network must select the appropriate bin in the light space to understand the light source configuration from the input image, the estimated normal map, and the albedos.

\subsection{Network Training}
We use the standard DiLiGenT benchmark dataset \cite{shi2016benchmark} having the $10$ objects imaged under $96$ different light directions with complex non-Lambertian surfaces. We implement DeepPS2 in Pytorch \cite{paszke2017automatic} with Adam optimizer \cite{kingma2014adam} and initial learning rate of $1 \times 10^{-4}$ for $25$ epochs and batch size $32$ on NVIDIA RTX $5000$ GPU. The learning rate is reduced to half after every $5$ epochs. It is observed that if the object under consideration has relatively simple reflectance properties, even a randomly initialized network trained with the image reconstruction loss can lead to good solutions. However, for complex scenes, it is better to warm up the network by initializing the weights through weak supervision only at the early stages of training \cite{taniai2018neural,kaya2021uncalibrated}. In our case, we perform this warming up for normal, albedo, and lighting estimation through weak supervision using $L_{1}$-loss ($\mathcal{L}_{L_{1}}$), $L_{2}$-loss ($\mathcal{L}_{L_{2}}$), and the perceptual loss ($\mathcal{L}_{perp}$) for first $2000$ iterations, as described in Section \ref{sec:loss}. For weak supervision, we randomly sample $10$ images (preferably, each one from a different lighting bin) and estimate the normal map using the least-squares formulation \cite{woodham1980photometric}, as per Equation \ref{eq:weak_sup}.
\begin{equation}
    \centering
    \hat{N'} = L^{-1}I
    \label{eq:weak_sup}
\end{equation}
It is important to note that the lighting directions in $L$ are from the discretized light space setting, where we compute the lighting direction as the one pointing towards the center of the selected bin. Since we have the images, the normal map $\hat{N'}$, and the discretized lightings $L$, we compute the \textit{diffuse shading} ($\boldsymbol{n}^{T}\boldsymbol{\ell}$) and \textit{specular highlights} (regions where $\boldsymbol{n}$ is close to the half-angle $\boldsymbol{h}$ of $\boldsymbol{\ell}$ and viewing direction $\boldsymbol{v} = [0,0,1]^{T}$). Once we have the shadings (diffuse and specular), we compute the albedos ($\hat{A'}$) to use them for weak supervision since an image is the product of the albedo and the shading.

\subsection{Loss Functions}\label{sec:loss}
In this section, we describe the loss function used for training the entire framework. Equation \ref{eq:loss_tot} describes the combination of $L_{1}$-loss and the perceptual loss $\mathcal{L}_{perp}$ used for both image reconstruction and relighting.
\begin{equation}
    \centering
    \mathcal{L}_{T}(X, \hat{X}) = \lambda_{1}\mathcal{L}_{1}(X, \hat{X}) + \lambda_{2}\mathcal{L}_{2}(X, \hat{X}) +  \lambda_{perp}\mathcal{L}_{perp}(X, \hat{X})
     \label{eq:loss_tot}
\end{equation}
Here,
\begin{equation*}
    \centering
    \mathcal{L}_{1}(X, \hat{X}) = \hspace{0.5em}\parallel X - \hat{X} \parallel_{1}
\end{equation*}
\begin{equation*}
    \centering
    \mathcal{L}_{2}(X, \hat{X}) = \hspace{0.5em}\parallel X - \hat{X} \parallel_{2}^{2}
\end{equation*}
\begin{equation}
    \centering
    \mathcal{L}_{perp}(X,\hat{X}) = \frac{1}{WHC}\sum_{x=1}^{W}\sum_{y=1}^{H}\sum_{z=1}^{C} \parallel \phi(X)_{x,y,z} - \phi(\hat{X})_{x,y,z} \parallel_{1}
\end{equation}
Here, $\phi$ is the output of VGG-19 \cite{simonyan2014very} network and $W$, $H$, $C$ are the width, height, and depth of the extracted feature $\phi$, respectively. $\lambda_{1} = \lambda_{2} = 0.5$ and $\lambda_{perp} = 1.0$.

\textbf{Weak Supervision.} We use the $\mathcal{L}_{T}$ and the standard cross-entropy loss to provide weak supervision (for first $2000$ iterations) for albedos and lightings, respectively. However, for surface normals, we use Equation \ref{eq:weak_sup_norm}. \begin{equation}
    \centering
    \mathcal{L}_{norm}(\hat{N}, \hat{N'}) = \frac{1}{M}\sum_{p} \parallel \hat{N}_{p} - \hat{N'}_{p} \parallel_{2}^{2} 
    \label{eq:weak_sup_norm}
\end{equation}
\vspace{-1cm}
\begin{figure}[h]
\centering
\includegraphics[width=0.9\linewidth]{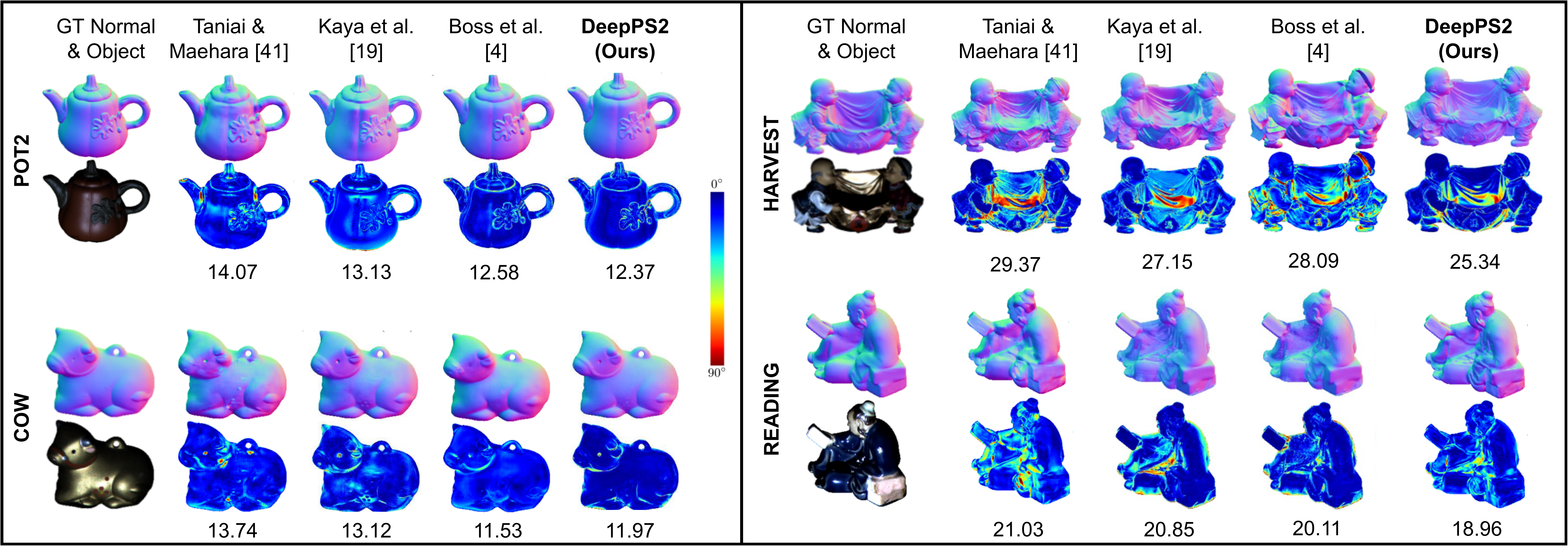}
\caption{Surface normal maps obtained using a randomly chosen input image pair}
\label{fig:res_qual_norm}
\end{figure}
\vspace{-1cm}
\section{Experimental Results}
In this section, we show the qualitative and quantitative comparison of the DeepPS2 with several baseline approaches. The classical methods \cite{queau2017photometric,mecca2011unambiguous} have provided the numerical resolution to the underlying ambiguities in PS2. However, the code and results on the DiLiGenT benchmark are not available for comparison. Moreover, since deep learning-based methods have significantly outperformed the traditional photometric stereo methods (even in handling ambiguities), we resort to comparing our work only with the state-of-the-art deep learning-based methods such as UPS-FCN \cite{chen2018ps}, SDPS-Net \cite{chen2019self}, IRPS \cite{taniai2018neural}, Kaya \emph{et al.} \cite{kaya2021uncalibrated}, Lichy \emph{et al.} \cite{lichy2021shape}, and Boss \emph{et al.} \cite{boss2020two}. They have been chosen carefully as they can be modified to align with our problem setting by re-training them with two images as input for a fair comparison.

\begin{figure}[h]
\centering
\includegraphics[width=0.8\linewidth]{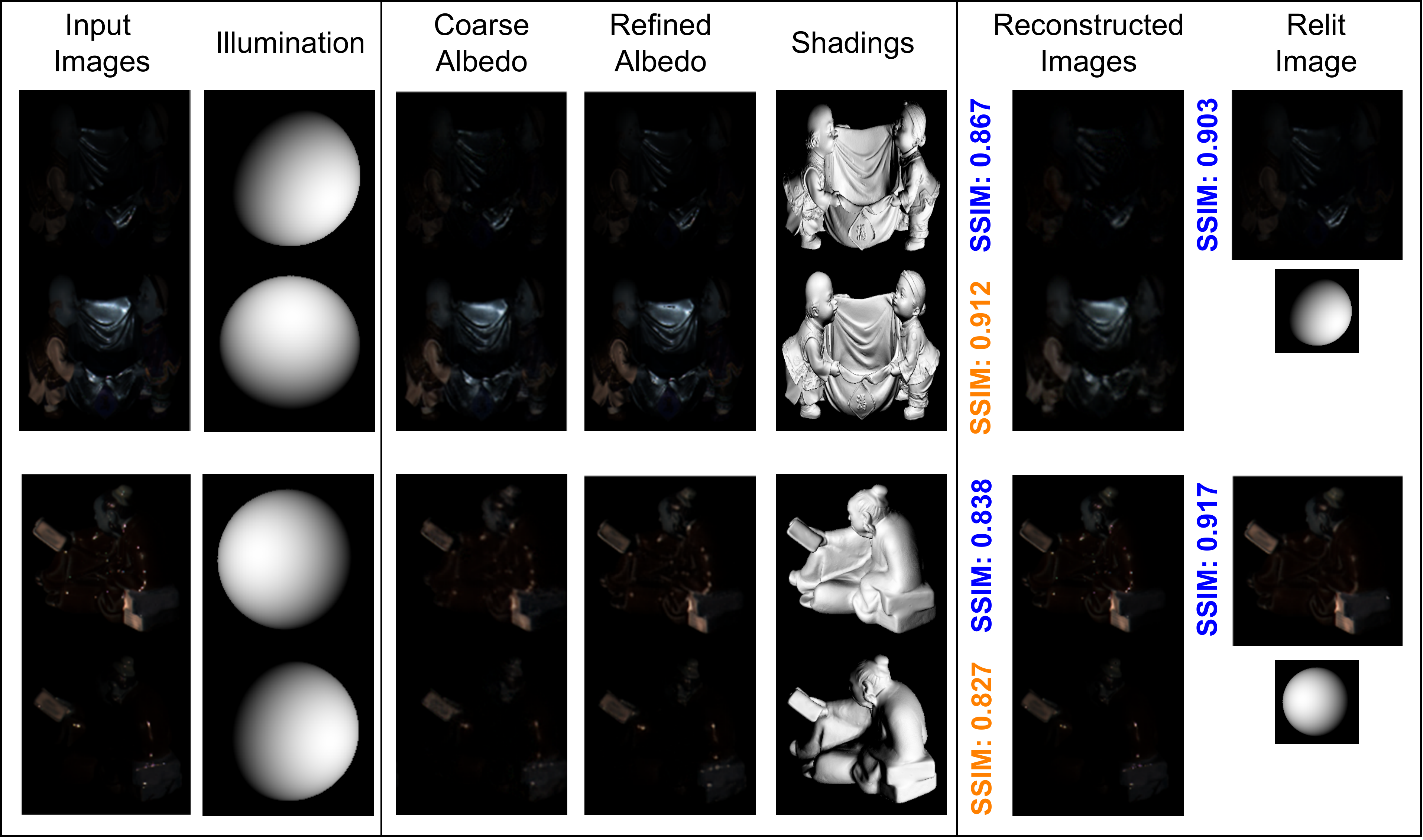}
\caption{Inverse rendering results on HARVEST and READING objects. The reconstruction and relighting module yield the SSIM of 0.837 and 0.779, respectively, when averaged over all the objects on the DiLiGenT Benchmark}
\label{fig:res_rel}
\end{figure}

\textbf{Results on Normal Estimation.} Table \ref{tab:quant_norm_estimation} shows a quantitative comparison of the proposed framework with the other deep learning-based methods. All the methods have been trained with two images as input, and the Mean Angular Error (MAE) is reported to quantify the surface normal estimation. Since IRPS \cite{taniai2018neural} is designed to take two images (one with frontal flash), we evaluate it using pairs of images where one image is lit frontally i.e., from the bin corresponding to $\theta = 0^{\circ}$ and $\phi = 90^{\circ}$. From Table \ref{tab:quant_norm_estimation}, we observe that the proposed DeepPS2 obtains the best average MAE value and best (or at least second best) individual scores for eight different objects (except POT1 and BEAR). Even though our framework performs best in the calibrated setting, it outperforms the other baselines under the uncalibrated setting as well. Furthermore, even with no ground truth supervision, our method outperforms other supervised (row 1-6) and self-supervised (row 7-8) methods. To appreciate the results qualitatively, we show a visual comparison of READING, HARVEST, COW, and POT2 with the self-supervised baselines \cite{taniai2018neural,kaya2021uncalibrated}, and a two-image based supervised method \cite{boss2020two} in Fig. \ref{fig:res_qual_norm}. Interestingly, DeepPS2 performs the best on objects like HARVEST and READING, having complex shadows and inter-reflections with spatially-varying material.

\setlength{\tabcolsep}{4pt}
\begin{table}[h]
\centering
\caption{Mean Angular Error (MAE) over 10 randomly chosen image pairs per object from the DiLiGenT Benchmark \cite{shi2016benchmark}. 
\textcolor{Green}{\textbf{GREEN}} and
\textcolor{Yellow}{\textbf{YELLOW}} coloured cells indicate the best and the second best performing methods, respectively. Rows 1-6 and 7-8 correspond to supervised and self-supervised approaches, respectively}
\label{tab:quant_norm_estimation}
\resizebox{\textwidth}{!}{%
\begin{tabular}{c|c|c|c|c|c|c|c|c|c|c|c|c}
\hline
\begin{tabular}[c]{@{}c@{}}Type of \\ Method\end{tabular} & \begin{tabular}[c]{@{}c@{}}Objects $\rightarrow $ \\ Method  $\downarrow$ \end{tabular} & Ball & Cat & Pot1 & Bear & Pot2 & Buddha & Goblet & Reading & Cow & Harvest & Average \\ \hline
Calibrated & PS-FCN \cite{chen2018ps} & 6.41 & 20.04 & 19.67 & 16.95 & 21.12 & 23.04 & 24.81 & 29.93 & 17.23 & 34.68 & 21.38 $\pm$ 2.05 \\
Uncalibrated & UPS-FCN \cite{chen2018ps} & 9.71 & 18.97 & 17.85 & 15.12 & 18.62 & 19.77 & 22.14 & 27.36 & 14.83 & 31.25 & 19.56  $\pm$  1.58 \\
Calibrated & SDPS-Net \cite{chen2019self} & 7.97 & 19.88 & 18.12 & 12.51 & 18.25 & 25.12 & 26.36 & 27.47 & 15.21 & 30.59 & 20.14  $\pm$  1.17\\ 
Uncalibrated & SDPS-Net \cite{chen2019self} & 7.81 & 21.74 & 19.73 & 13.25 & 20.47 & 27.81 & 29.66 & 31.12 & 18.94 & 34.14 & 22.6  $\pm$  1.02 \\
Uncalibrated & Boss \emph{et al.} \cite{boss2020two} & 7.71 & 14.81 & 
\cellcolor{Yellow}10.17 & \cellcolor{Green}8.01 & 12.89 & 
15.98 & 18.18 & 21.54 & 11.96 & 27.36 & 14.85  $\pm$  0.98  \\
Uncalibrated & Lichy\emph{et al.} \cite{lichy2021shape} & 7.42 & 20.34 & 11.87 & 9.94 & \cellcolor{Green}11.12 & 18.75 & 19.38 & 21.51 & 12.93 & 29.52 & 16.27  $\pm$  1.01 \\ \hline
Calibrated & Taniai \& Maehara \cite{taniai2018neural}& 7.03 & 10.02 & 11.62 & 8.74 & 12.58 & 18.25 & 16.85 & 21.31 & 14.97 & 28.89 & 15.03  $\pm$  0.96 \\
Uncalibrated & Kaya \emph{et al.} \cite{kaya2021uncalibrated} & 6.97 & \cellcolor{Green}9.57 & \cellcolor{Green}10.14 & \cellcolor{Yellow}8.69 & 13.81 & 17.57 & 15.93 & 21.87 & 14.81 & 28.72 & 14.81  $\pm$  0.89\\ \hline
Calibrated & DeepPS2 (Ours) & \cellcolor{Green}6.17 & \cellcolor{Yellow}9.62 & 10.35 & 8.87 & 12.78 & \cellcolor{Yellow}14.78 & \cellcolor{Green}13.29 & \cellcolor{Green}18.34 & \cellcolor{Green}10.13 & \cellcolor{Green}25.18 & \cellcolor{Green}12.95  $\pm$  0.64 \\ 
Uncalibrated & DeepPS2 (Ours) & \cellcolor{Yellow}6.28 & 9.87 & 10.73 & 9.67 & \cellcolor{Yellow}12.09 & \cellcolor{Green}14.51 & \cellcolor{Yellow}14.22 & \cellcolor{Yellow}19.94 & \cellcolor{Yellow}11.08 & \cellcolor{Yellow}26.06 & \cellcolor{Yellow}13.44  $\pm$  0.67\\\hline
\end{tabular}}
\end{table}
\setlength{\tabcolsep}{1.4pt}
\vspace{-0.5cm}
\textbf{Results on Albedo Estimation.} In Fig. \ref{fig:res_rel}, we present a qualitative assessment of the albedos obtained using our method. We observe that the learned albedos are able to handle the complex shadows and specular highlights, especially after refinement using the estimated lightings. 

\textbf{Results on Lighting Estimation.} The goal of discretized lighting is to remove the network's dependence on precise lighting calibration. Therefore, we attempt to model the illumination using the weakly calibrated lighting directions such as front, front-right/left, top, top-right/left, bottom, bottom-right/left, etc. Given that the light space discretization yields an MAE of $18^{\circ}$ numerically, we intend to establish that the network may not need precise calibration at all times. A rough and/or abstract understanding of lighting directions should help guide the network towards realistic shape estimation. To better evaluate the performance of the \textit{illumination module}, we visualize the learned illumination over a sphere in Fig. \ref{fig:res_rel}. It is observed that the illumination module captures the distribution of light sources essential for modeling the complex specularities in the refined albedos at the later stage.

\textbf{Results on Image Relighting and Reconstruction.} We report the widely used Structural Similarity Index (SSIM) \cite{wang2003multiscale} to quantify the quality of the reconstructed and relight images. However, these results are best appreciated visually. Therefore, we use Fig. \ref{fig:res_rel} to show the quality of the generated images. The quality of the results establishes that our inverse rendering results are sufficiently stable for realistic relighting and reconstruction.

\subsection{Ablation Studies}
In this section, we discuss several design choices in DeepPS2 under different experimental settings.

\textbf{Ablation 1: \textit{What if we do not include lighting estimation in the framework?}} We attempt to understand the effect of including the lighting information explicitly in the surface normal estimation through such an inverse rendering-based framework. In Table \ref{tab:quant_ablation}, comparing the experiment IDs 1 and 2, we observe that lighting estimation is crucial for the task at hand. This observation is in line with the classical rendering equation that requires lighting directions to understand the reflectance properties and shadows on the surface.
Further, we intended to know the deviation in MAE for surface normal estimation when actual lightings (calibrated setting) are used. Although the network performs better under the calibrated setting (see Table \ref{tab:quant_norm_estimation}), the error difference is not very large ($0.49$ units). This supports our idea of using weaker calibrations for surface normal estimation under distant lightings.
\vspace{-0.5cm}
\setlength{\tabcolsep}{4pt}
\begin{table}[h]
\centering
\caption{Quantitative comparison of various design choices. LE: Lighting Estimation, AR: Albedo Refinement, PE: Positional Encoding, and IR: Image Relighting. Experiments IDs 1-6 include warm-up}
\label{tab:quant_ablation}
\resizebox{\textwidth}{!}{%
\begin{tabular}{c|cccc|cccccccccc|c}\hline
ID & LE & AR & PE & IR & Ball & Cat & Pot1 & Bear & Pot2 & Buddha & Goblet & Reading & Cow & Harvest & Average \\\hline
1 & \ding{55} & \ding{55} & \ding{55} & \ding{55} & 9.87 & 36.55 & 19.39 & 12.42 & 14.52 & 13.19 & 20.57 & 58.96 & 19.75 & 55.51 & 26.07 \\
2 & \ding{51} & \ding{55} & \ding{55} & \ding{55} & 9.32 & 15.62 & 16.41 & 10.96 & 15.77 & 19.93 & 18.37 & 32.34 & 16.17 & 30.26 & 18.51 \\
3 & \ding{51} & \ding{51} & \ding{55} & \ding{55} & 7.37 & 15.64 & 10.58 & 9.37 & 14.72 & 15.06 & 18.1 & 23.78 & 16.31 & 27.17 & 15.85 \\
4 & \ding{51} & \ding{51} & \ding{51} & \ding{55} & 6.88 & 12.16 & 11.13 & 9.79 & 15.11 & 14.89 & 16.07 & 20.46 & 11.85 & 27.22 & 14.55 \\
5 & \ding{51} & \ding{51} & \ding{51} & \ding{51} & 6.28 & 9.87 & 10.73 & 9.67 & 12.09 & 14.51 & 14.22 & 19.94 & 11.08 & 26.06 & 13.44 \\
6 & \multicolumn{4}{c|}{frontally-lit image} & 6.74 & 9.38 & 10.13 & 9.08 & 13.18 & 14.58 & 14.63 & 17.84 & 11.98 & 24.87 & 13.24 \\ \hline
7 & \multicolumn{4}{c|}{w/o warm-up} & 12.43  & 25.01 & 22.82 & 15.44 & 20.57 & 25.76 & 29.16 & 52.16 & 25.53 & 44.45 & 27.33 \\\hline
8 & \multicolumn{4}{c|}{fully supervised} & 5.14 & 8.97 & 10.28 & 8.92 & 9.89 & 12.76 & 12.38 & 18.52 & 9.81 & 23.22 & 11.98 \\ \hline 
\end{tabular}%
}
\end{table}
\setlength{\tabcolsep}{1.4pt}

\textbf{Ablation 2: \textit{Effect of discretizing the light space on normal estimation.}}
Fig. \ref{fig:light_space} (b) shows the effect of a different number of bins on the MAE evaluated over the DiLiGenT benchmark. We resort to choosing $K=25$ bins as the reduction in the MAE plateaus (roughly) after that point. Further, the light space discretization not only reduces the computational overhead but also helps the network understand the lighting dynamics more holistically. This is evident from the MAE reported in Table \ref{tab:quant_norm_estimation} and quality of the refined albedos in Fig. \ref{fig:res_rel}.

\textbf{Ablation 3: \textit{Do albedo refinement and image relighting help in modeling the illumination?}} Qualitative results in Fig. \ref{fig:res_rel} show how well the refined albedos capture the specularities on the surface. Table \ref{tab:quant_ablation} (IDs 2 and 3) shows the performance improvement by including the \textit{albedo refinement module}. The explicit specularity modeling is observed to produce realistic albedos. The performance is further enhanced through the use of positional encoding (Table \ref{tab:quant_ablation} ID 4) as it helps the module to better capture the high-frequency characteristics in the refined albedo. Finally, the inclusion of the \textit{image relighting module} further reduces the MAE (Table \ref{tab:quant_ablation} ID 5). Since the relighting module is solely driven by the estimated lightings, relighting helps in obtaining better surface normal estimates through better lighting estimation as an additional task.

\textbf{Ablation 4: \textit{What is the effect of warming up the network with weak supervision at the early stages of training?}} We also consider understanding the effect of weak supervision during the early stage warm-up. Table \ref{tab:quant_ablation} (IDs 5 and 7) clearly establishes the benefit of warming-up. Fig. \ref{fig:light_space} (c) shows the the convergence with and without the warm-up. Clearly, an early-stage warm-up provides stable and faster convergence as the outliers in the images are excluded at the early stages during weak supervision.

\textbf{Ablation 5: \textit{What if the lighting directions of one image at the input is known?}} We evaluate an interesting and practical case where one of the two input images is captured with collocated light source and camera i.e., $\boldsymbol{\ell} = \boldsymbol{v} = [0,0,1]^{T}$. Since the lighting direction is known, we provide (auxiliary) supervision to the illumination module to obtain a better lighting estimate for the other image. Table \ref{tab:quant_ablation} (ID 6) shows the results obtained over image pairs having one image sampled from the frontal lighting bin i.e. $\theta = 0^{\circ}, \phi=90^{\circ}$. Under this setting, the method performs better than the completely self-supervised version because frontally-lit (flashed) images offer a better understanding of specularities on complex surfaces. Finally, we also show the performance of DeepPS2 under a fully supervised setting (Table \ref{tab:quant_ablation} (ID 8)) to establish the upper bound of DeepPS2.

\section{Conclusion}

In this work, we address the PS2 problem (photometric stereo with two images) using a self-supervised deep learning framework called DeepPS2. In addition to surface normals, the proposed method also estimates albedos and lightings and performs image relighting, all without any ground truth supervision. Interestingly, we demonstrate that weakly calibrated lightings can be enough for the network to learn the underlying shape of an object. In conjunction with image reconstruction, image relighting helps in better lighting estimation. While other uncalibrated methods have used ground truth supervision for learning to estimate lightings, we do so entirely in a self-supervised manner. To the best of our knowledge, we are the first to address photometric stereo using two images in a deep learning setting. 

% Owing to the applicability of the PS2 setting, we would like to extend this work for modelling outdoor illumination and non-rigid shape reconstruction in the near future. 

% \clearpage\mbox{}Page \thepage\ of the manuscript.
% \clearpage\mbox{}Page \thepage\ of the manuscript.

% This is the last page of the manuscript.
% Now we have reached the maximum size of the ECCV 2022 submission (excluding references).
% References should start immediately after the main text, but can continue on p.15 if needed.

\clearpage
\bibliographystyle{splncs04}
\bibliography{main}

\begin{thebibliography}{10}
\providecommand{\url}[1]{\texttt{#1}}
\providecommand{\urlprefix}{URL }
\providecommand{\doi}[1]{https://doi.org/#1}

\bibitem{abrams2012heliometric}
Abrams, A., Hawley, C., Pless, R.: Heliometric stereo: Shape from sun position.
  In: European conference on computer vision. pp. 357--370. Springer (2012)

\bibitem{ackermann2012photometric}
Ackermann, J., Langguth, F., Fuhrmann, S., Goesele, M.: Photometric stereo for
  outdoor webcams. In: 2012 IEEE conference on computer vision and pattern
  recognition. pp. 262--269. IEEE (2012)

\bibitem{belhumeur1999bas}
Belhumeur, P.N., Kriegman, D.J., Yuille, A.L.: The bas-relief ambiguity.
  International journal of computer vision  \textbf{35}(1),  33--44 (1999)

\bibitem{boss2020two}
Boss, M., Jampani, V., Kim, K., Lensch, H., Kautz, J.: Two-shot
  spatially-varying brdf and shape estimation. In: Proceedings of the IEEE/CVF
  Conference on Computer Vision and Pattern Recognition. pp. 3982--3991 (2020)

\bibitem{boykov2001fast}
Boykov, Y., Veksler, O., Zabih, R.: Fast approximate energy minimization via
  graph cuts. IEEE Transactions on pattern analysis and machine intelligence
  \textbf{23}(11),  1222--1239 (2001)

\bibitem{burley2012physically}
Burley, B., Studios, W.D.A.: Physically-based shading at disney. In: ACM
  SIGGRAPH. vol.~2012, pp.~1--7. vol. 2012 (2012)

\bibitem{chen2019self}
Chen, G., Han, K., Shi, B., Matsushita, Y., Wong, K.Y.K.: Self-calibrating deep
  photometric stereo networks. In: Proceedings of the IEEE Conference on
  Computer Vision and Pattern Recognition. pp. 8739--8747 (2019)

\bibitem{chen2020deep}
Chen, G., Han, K., Shi, B., Matsushita, Y., Wong, K.Y.K.: Deep photometric
  stereo for non-lambertian surfaces. IEEE Transactions on Pattern Analysis and
  Machine Intelligence  (2020)

\bibitem{chen2018ps}
Chen, G., Han, K., Wong, K.Y.K.: Ps-fcn: A flexible learning framework for
  photometric stereo. In: Proceedings of the European Conference on Computer
  Vision (ECCV). pp. 3--18 (2018)

\bibitem{chen2020learned}
Chen, G., Waechter, M., Shi, B., Wong, K.Y.K., Matsushita, Y.: What is learned
  in deep uncalibrated photometric stereo? In: European Conference on Computer
  Vision. pp. 745--762. Springer (2020)

\bibitem{furukawa2009accurate}
Furukawa, Y., Ponce, J.: Accurate, dense, and robust multiview stereopsis. IEEE
  transactions on pattern analysis and machine intelligence  \textbf{32}(8),
  1362--1376 (2009)

\bibitem{hernandez2007non}
Hern{\'a}ndez, C., Vogiatzis, G., Brostow, G.J., Stenger, B., Cipolla, R.:
  Non-rigid photometric stereo with colored lights. In: 2007 IEEE 11th
  International Conference on Computer Vision. pp.~1--8. IEEE (2007)

\bibitem{hernandez2010overcoming}
Hern{\'a}ndez, C., Vogiatzis, G., Cipolla, R.: Overcoming shadows in 3-source
  photometric stereo. IEEE Transactions on Pattern Analysis and Machine
  Intelligence  \textbf{33}(2),  419--426 (2010)

\bibitem{horn1970shape}
Horn, B.K.: Shape from shading: A method for obtaining the shape of a smooth
  opaque object from one view  (1970)

\bibitem{ikeda2003robust}
Ikeda, O.: A robust shape-from-shading algorithm using two images and control
  of boundary conditions. In: Proceedings 2003 International Conference on
  Image Processing (Cat. No. 03CH37429). vol.~1, pp. I--405. IEEE (2003)

\bibitem{ikehata2018cnn}
Ikehata, S.: Cnn-ps: Cnn-based photometric stereo for general non-convex
  surfaces. In: Proceedings of the European conference on computer vision
  (ECCV). pp. 3--18 (2018)

\bibitem{ikeuchi1981numerical}
Ikeuchi, K., Horn, B.K.: Numerical shape from shading and occluding boundaries.
  Artificial intelligence  \textbf{17}(1-3),  141--184 (1981)

\bibitem{jung2015one}
Jung, J., Lee, J.Y., So~Kweon, I.: One-day outdoor photometric stereo via
  skylight estimation. In: Proceedings of the IEEE Conference on Computer
  Vision and Pattern Recognition. pp. 4521--4529 (2015)

\bibitem{kaya2021uncalibrated}
Kaya, B., Kumar, S., Oliveira, C., Ferrari, V., Van~Gool, L.: Uncalibrated
  neural inverse rendering for photometric stereo of general surfaces. In:
  Proceedings of the IEEE/CVF Conference on Computer Vision and Pattern
  Recognition. pp. 3804--3814 (2021)

\bibitem{kendall2017end}
Kendall, A., Martirosyan, H., Dasgupta, S., Henry, P., Kennedy, R., Bachrach,
  A., Bry, A.: End-to-end learning of geometry and context for deep stereo
  regression. In: Proceedings of the IEEE international conference on computer
  vision. pp. 66--75 (2017)

\bibitem{kingma2014adam}
Kingma, D.P., Ba, J.: Adam: A method for stochastic optimization. arXiv
  preprint arXiv:1412.6980  (2014)

\bibitem{kozera1992shape}
Kozera, R.: On shape recovery from two shading patterns. International Journal
  of Pattern Recognition and Artificial Intelligence  \textbf{6}(04),  673--698
  (1992)

\bibitem{kumar2019jumping}
Kumar, S.: Jumping manifolds: Geometry aware dense non-rigid structure from
  motion. In: Proceedings of the IEEE/CVF Conference on Computer Vision and
  Pattern Recognition. pp. 5346--5355 (2019)

\bibitem{kumar2017monocular}
Kumar, S., Dai, Y., Li, H.: Monocular dense 3d reconstruction of a complex
  dynamic scene from two perspective frames. In: Proceedings of the IEEE
  international conference on computer vision. pp. 4649--4657 (2017)

\bibitem{kumar2019superpixel}
Kumar, S., Dai, Y., Li, H.: Superpixel soup: Monocular dense 3d reconstruction
  of a complex dynamic scene. IEEE transactions on pattern analysis and machine
  intelligence  \textbf{43}(5),  1705--1717 (2019)

\bibitem{li2019learning}
Li, J., Robles-Kelly, A., You, S., Matsushita, Y.: Learning to minify
  photometric stereo. In: Proceedings of the IEEE/CVF Conference on Computer
  Vision and Pattern Recognition. pp. 7568--7576 (2019)

\bibitem{lichy2021shape}
Lichy, D., Wu, J., Sengupta, S., Jacobs, D.W.: Shape and material capture at
  home. In: Proceedings of the IEEE/CVF Conference on Computer Vision and
  Pattern Recognition. pp. 6123--6133 (2021)

\bibitem{mecca2011unambiguous}
Mecca, R., Durou, J.D.: Unambiguous photometric stereo using two images. In:
  International Conference on Image Analysis and Processing. pp. 286--295.
  Springer (2011)

\bibitem{onn1990integrability}
Onn, R., Bruckstein, A.: Integrability disambiguates surface recovery in
  two-image photometric stereo. International Journal of Computer Vision
  \textbf{5}(1),  105--113 (1990)

\bibitem{pacanowski2012rational}
Pacanowski, R., Celis, O.S., Schlick, C., Granier, X., Poulin, P., Cuyt, A.:
  Rational brdf. IEEE transactions on visualization and computer graphics
  \textbf{18}(11),  1824--1835 (2012)

\bibitem{paszke2017automatic}
Paszke, A., Gross, S., Chintala, S., Chanan, G., Yang, E., DeVito, Z., Lin, Z.,
  Desmaison, A., Antiga, L., Lerer, A.: Automatic differentiation in pytorch
  (2017)

\bibitem{prados2005shape}
Prados, E., Faugeras, O.: Shape from shading: a well-posed problem? In: 2005
  IEEE computer society conference on computer vision and pattern recognition
  (CVPR'05). vol.~2, pp. 870--877. IEEE (2005)

\bibitem{queau2017photometric}
Qu{\'e}au, Y., Mecca, R., Durou, J.D., Descombes, X.: Photometric stereo with
  only two images: A theoretical study and numerical resolution. Image and
  Vision Computing  \textbf{57},  175--191 (2017)

\bibitem{rusinkiewicz1998new}
Rusinkiewicz, S.M.: A new change of variables for efficient brdf
  representation. In: Eurographics Workshop on Rendering Techniques. pp.
  11--22. Springer (1998)

\bibitem{santo2017deep}
Santo, H., Samejima, M., Sugano, Y., Shi, B., Matsushita, Y.: Deep photometric
  stereo network. In: Proceedings of the IEEE international conference on
  computer vision workshops. pp. 501--509 (2017)

\bibitem{sato1995reflectance}
Sato, Y., Ikeuchi, K.: Reflectance analysis under solar illumination. In:
  Proceedings of the Workshop on Physics-Based Modeling in Computer Vision. pp.
  180--187. IEEE (1995)

\bibitem{schonberger2016structure}
Schonberger, J.L., Frahm, J.M.: Structure-from-motion revisited. In:
  Proceedings of the IEEE conference on computer vision and pattern
  recognition. pp. 4104--4113 (2016)

\bibitem{shi2016benchmark}
Shi, B., Wu, Z., Mo, Z., Duan, D., Yeung, S.K., Tan, P.: A benchmark dataset
  and evaluation for non-lambertian and uncalibrated photometric stereo. In:
  Proceedings of the IEEE Conference on Computer Vision and Pattern
  Recognition. pp. 3707--3716 (2016)

\bibitem{simonyan2014very}
Simonyan, K., Zisserman, A.: Very deep convolutional networks for large-scale
  image recognition. arXiv preprint arXiv:1409.1556  (2014)

\bibitem{taniai2018neural}
Taniai, T., Maehara, T.: Neural inverse rendering for general reflectance
  photometric stereo. In: International Conference on Machine Learning. pp.
  4857--4866. PMLR (2018)

\bibitem{taniai2017continuous}
Taniai, T., Matsushita, Y., Sato, Y., Naemura, T.: Continuous 3d label stereo
  matching using local expansion moves. IEEE transactions on pattern analysis
  and machine intelligence  \textbf{40}(11),  2725--2739 (2017)

\bibitem{tiwari2022lerps}
Tiwari, A., Raman, S.: Lerps: Lighting estimation and relighting for
  photometric stereo. In: ICASSP 2022-2022 IEEE International Conference on
  Acoustics, Speech and Signal Processing (ICASSP). pp. 2060--2064. IEEE (2022)

\bibitem{wang2020non}
Wang, X., Jian, Z., Ren, M.: Non-lambertian photometric stereo network based on
  inverse reflectance model with collocated light. IEEE Transactions on Image
  Processing  \textbf{29},  6032--6042 (2020)

\bibitem{wang2003multiscale}
Wang, Z., Simoncelli, E.P., Bovik, A.C.: Multiscale structural similarity for
  image quality assessment. In: The Thrity-Seventh Asilomar Conference on
  Signals, Systems \& Computers, 2003. vol.~2, pp. 1398--1402. Ieee (2003)

\bibitem{woodham1980photometric}
Woodham, R.J.: Photometric method for determining surface orientation from
  multiple images. Optical engineering  \textbf{19}(1),  139--144 (1980)

\bibitem{yang2017stacked}
Yang, J., Liu, Q., Zhang, K.: Stacked hourglass network for robust facial
  landmark localisation. In: Proceedings of the IEEE Conference on Computer
  Vision and Pattern Recognition Workshops. pp. 79--87 (2017)

\bibitem{yang1992two}
Yang, J., Ohnishi, N., Sugie, N.: Two-image photometric stereo method. In:
  Intelligent Robots and Computer Vision XI: Biological, Neural Net, and 3D
  Methods. vol.~1826, pp. 452--463. SPIE (1992)

\bibitem{yao2020gps}
Yao, Z., Li, K., Fu, Y., Hu, H., Shi, B.: Gps-net: Graph-based photometric
  stereo network. Advances in Neural Information Processing Systems
  \textbf{33},  10306--10316 (2020)

\bibitem{zheng2019spline}
Zheng, Q., Jia, Y., Shi, B., Jiang, X., Duan, L.Y., Kot, A.C.: Spline-net:
  Sparse photometric stereo through lighting interpolation and normal
  estimation networks. In: Proceedings of the IEEE/CVF International Conference
  on Computer Vision. pp. 8549--8558 (2019)

\end{thebibliography}

\newpage

\section{Supplementary Material}

Although the main paper is self-contained in terms of the main results, we believe that the supplementary material can be of help to understand the work in greater detail. Here, we describe the DeepPS2 architecture and remaining results on surface normal, albedo, shading, and illumination estimation. Also, we demonstrate qualitatively the results of image reconstruction and relighting on different objects from the DiLiGenT benchmark \cite{shi2016benchmark}. 

\subsection{DeepPS2 Architecture}
Table \ref{tab:net_arch} describes the detailed network architecture. The design of all the modules (except the \textit{illumination module}) is inspired by that of Hourglass networks \cite{yang2017stacked}.

\setlength{\tabcolsep}{4pt}
\begin{table}[]
\caption{Detailed network architecture of DeepPS2}
\label{tab:net_arch}
\centering
\resizebox{0.85\textwidth}{!}{%
\begin{tabular}{l|l}\hline
\multicolumn{1}{c|}{Module} & \multicolumn{1}{c}{Architecture} \\\hline
\multicolumn{1}{c|}{Encoder} & \begin{tabular}[c]{@{}l@{}}conv(k=6, p=2, s=2, cin = 7, cout = 32), BN, ReLU\\ conv(k=4, p=1, s=2, cin = 32, cout = 64), BN, ReLU\\ conv(k=4, p=1, s=2, cin = 64, cout = 128), BN, ReLU\\ conv(k=4, p=1, s=2, cin = 128, cout = 256), BN, ReLU\\ conv(k=4, p=1, s=2, cin=256, cout = 512), BN, ReLU\end{tabular} \\\hline
\multicolumn{1}{c|}{\begin{tabular}[c]{@{}c@{}}Decoder\\ (Normal and Albedo)\end{tabular}} & \begin{tabular}[c]{@{}l@{}}conv(k=4, p=1, s=2 cin = 512, cout = 256), BN, ReLU\\ conv(k=4, p=1, s=2, cin = 512, cout = 128), BN, ReLU\\ conv(k=4, p=1, s=2, cin = 256, cout = 64), BN, ReLU\\ conv(k=4, p=1, s=2, cin = 128, cout = 32), BN, ReLU\\ conv(k=4, p=1, s=2, cin = 64, cout = 64), BN, ReLU\\ Normal: conv(k=5, p=2, s=1, cin=64, cout = 3), Tanh\\ Albedo:  conv(k=5, p=2, s=1, cin=64, cout=6), Tanh\end{tabular} \\\hline
\multicolumn{1}{c|}{Illumination} & \begin{tabular}[c]{@{}l@{}}conv(k=3, p=0, s=1, cin = 9, cout = 64), BN, ReLU\\ conv(k=3, p=0, s=1, cin = 64, cout = 128), BN, ReLU\\ conv(k=3, p=0, s=1, cin = 128, cout = 256), BN, ReLU\end{tabular} \\
 & \begin{tabular}[c]{@{}l@{}}\hline\textbf{Regress $\theta$:}\\ Linear(256, 256), ReLU, Dropout(0.25)\\ Linear(256,64), ReLU, DropOut(0.25)\\ Linear(64, 5)\\\hline\end{tabular} \\
 & \begin{tabular}[c]{@{}l@{}}\textbf{Regress $\phi$:}\\ Linear(256, 256), ReLU, Dropout(0.25)\\ Linear(256,64), ReLU, DropOut(0.25)\\ Linear(64, 5)\end{tabular} \\\hline
Albedo Refinement & \begin{tabular}[c]{@{}l@{}}conv(k=6, p=2, s=2, cin = 44, cout = 128), BN, ReLU\\ conv(k=4, p=1, s=2, cin = 128, cout = 128), BN, ReLU\\ conv(k=4, p=1, s=2, cin = 128, cout = 256), BN, ReLU\\ conv(k=4, p=1, s=2, cin = 256, cout = 128), BN, ReLU\\ conv(k=4, p=1, s=2, cin =256, cout = 128), BN, ReLU\\ conv(k=4, p=1, s=2, cin =256, cout = 64), BN, ReLU\\ conv(k=5, p=2, s=1, cin =64, cout = 6), Tanh\end{tabular} \\ \hline
Image Reconstruction & \begin{tabular}[c]{@{}l@{}}conv(k=6, p=2, s=2, cin = 15, cout = 64), BN, ReLU\\ conv(k=4, p=1, s=2, cin = 64, cout = 128), BN, ReLU\\ conv(k=4, p=1, s=2, cin = 128, cout = 128), BN, ReLU\\ conv(k=4, p=1, s=2, cin = 128, cout = 256), BN, ReLU\\ conv(k=4, p=1, s=2, cin=256, cout = 128), BN, ReLU\\ conv(k=4, p=1, s=2, cin=256, cout = 128), BN, ReLU\\ conv(k=4, p=1, s=2, cin=256, cout=64), BN, ReLU\\ conv(k=4, p=1, s=2, cin=128, cout=64), BN, ReLU\\ conv(k=5, p=2, s=1, cin=64, cout=6), Tanh\end{tabular} \\ \hline
\multicolumn{1}{c|}{Image Relighting} & \begin{tabular}[c]{@{}l@{}}conv(k=6, p=2, s=2, cin = 7, cout = 64), BN, ReLU\\ conv(k=4, p=1, s=2, cin = 64, cout = 128), BN, ReLU\\ conv(k=4, p=1, s=2, cin = 128, cout = 128), BN, ReLU\\ conv(k=4, p=1, s=2, cin = 128, cout = 256), BN, ReLU\\ lighting feature(256)\\ conv(k=4, p=1, s=2, cin=256, cout = 128), BN, ReLU\\ conv(k=4, p=1, s=2, cin=256, cout = 128), BN, ReLU\\ conv(k=4, p=1, s=2, cin=256, cout=64), BN, ReLU\\ conv(k=4, p=1, s=2, cin=128, cout=64), BN, ReLU\\ conv(k=5, p=2, s=1, cin=64, cout=6), Tanh\end{tabular} \\
 & \begin{tabular}[c]{@{}l@{}}\hline \textbf{Lighting Feature}:\\ conv(k=1, p=0, s=1, cin=3, cout=64)\\ conv(k=1, p=0, s=1, cin=64, cout=128), BN, Upsample(2)\\ conv(k=3, p=1, s=1, cin=128, cout=128), BN, Upsample(2)\\ conv(k=3, p=1, s=1, cin=128, cout=256), BN, Upsample(2)\\ conv(k=3, p=1, s=1, cin=256, cout=256)\\\end{tabular}\\\hline
\end{tabular}%
}
\end{table}
\setlength{\tabcolsep}{1.4pt}

\subsection{Results on Normal Estimation}
\begin{figure}[h]
\centering
\includegraphics[width = 0.8\textwidth]{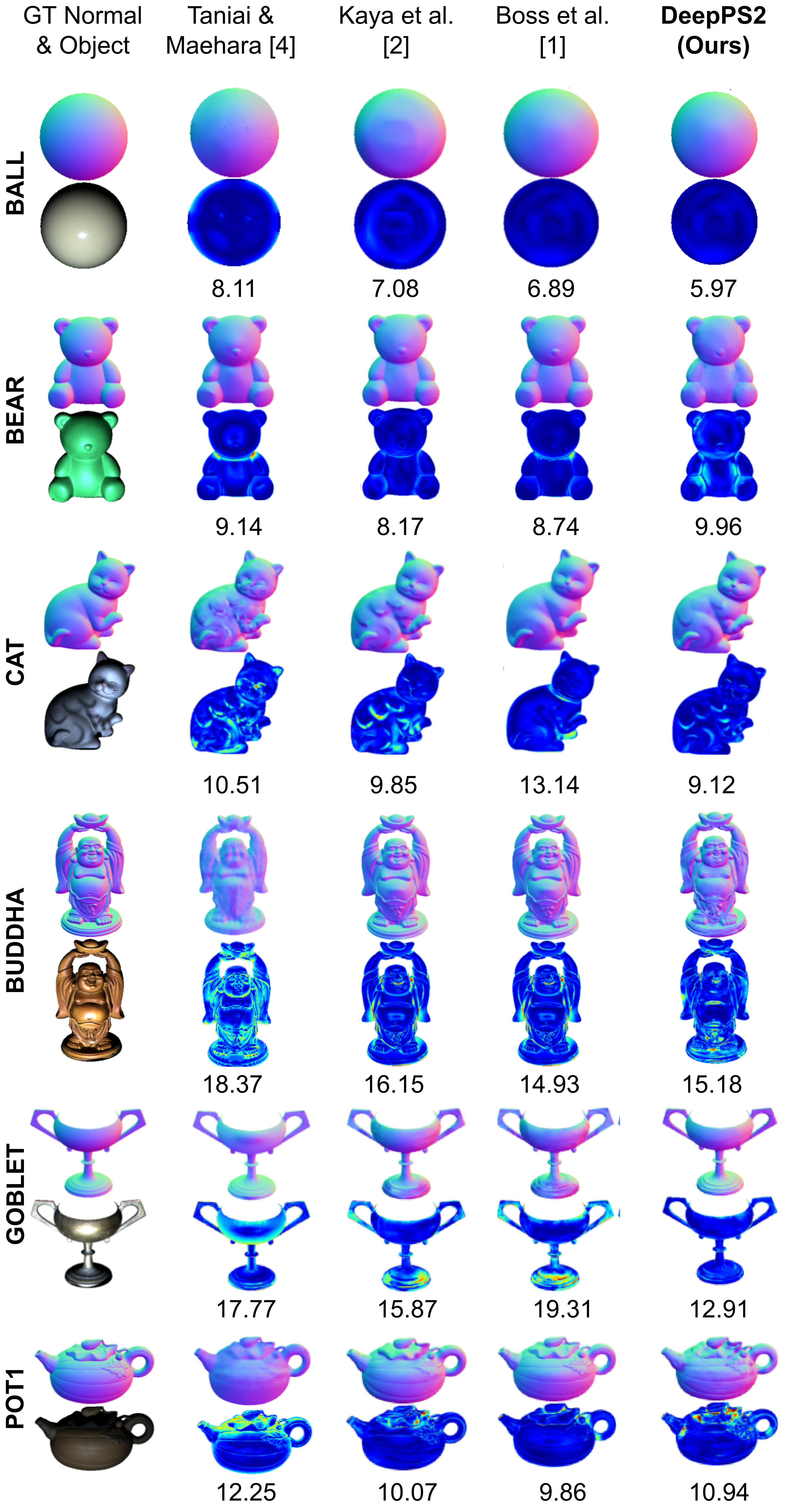}
\caption{Normal estimation results on remaining objects in the DiLiGenT benchmark }
\label{fig:res_ne_supp}
\end{figure}

Figure \ref{fig:res_ne_supp} shows the qualitative comparison of the surface normal maps obtained using DeepPS2 with other baselines \cite{taniai2017continuous,kaya2021uncalibrated,boss2020two} over the six remaining objects on the DiLiGenT benchmark dataset.

\subsection{Additional Inverse Rendering Results}

\begin{figure}[h]
\centering
\includegraphics[width = \textwidth]{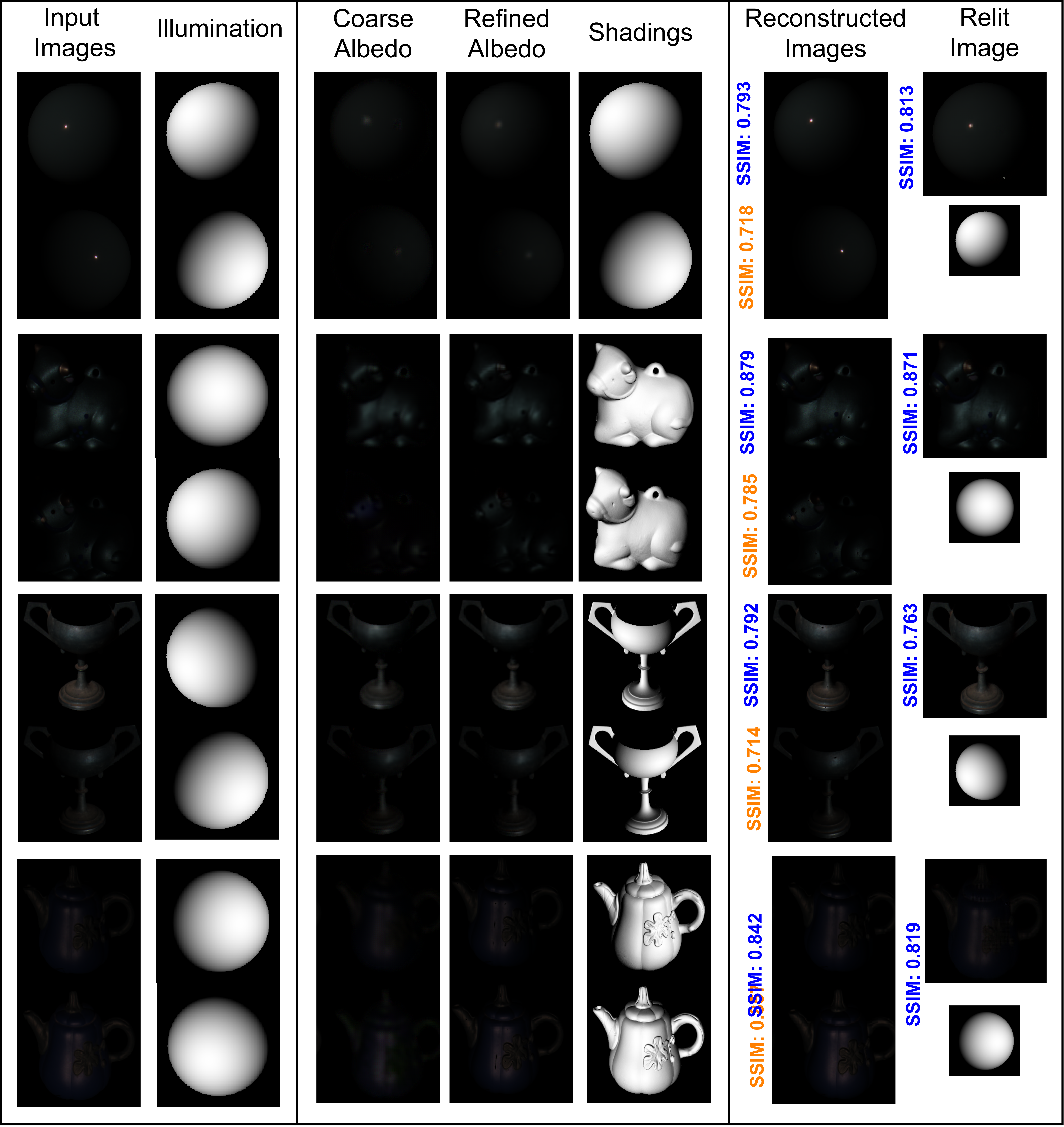}
\caption{Inverse rendering results on additional objects from DiLiGenT benchmark}
\label{fig:res_rel_supp}
\end{figure}

Figure \ref{fig:res_rel_supp} shows the qualitative comparison of the estimated illumination, albedo, and shading through DeepPS2. We also show the image reconstruction and relighting results along with the SSIM value. 

\clearpage
\end{document}